\documentclass[a4paper]{article}
\usepackage{multirow}
\usepackage{INTERSPEECH2021}
\usepackage{color}
\usepackage{todonotes}
\usepackage{url}

\title{Multilingual and code-switching ASR challenges for \\ low resource Indian languages}
\name{Anuj Diwan$^1$, Rakesh Vaideeswaran$^2$, Sanket Shah$^3$, Ankita Singh$^1$, Srinivasa Raghavan$^4$, Shreya Khare$^5$, Vinit Unni$^1$, Saurabh Vyas$^4$, Akash Rajpuria$^4$, Chiranjeevi Yarra$^6$, Ashish Mittal$^5$, Prasanta Kumar Ghosh$^2$, Preethi Jyothi$^1$, Kalika Bali$^3$, Vivek Seshadri$^3$, Sunayana Sitaram$^3$, Samarth Bharadwaj$^5$, Jai Nanavati$^4$, Raoul Nanavati$^4$, Karthik Sankaranarayanan$^5$,\thanks{Additional authors: Tejaswi Seeram and Basil Abraham, Microsoft Corporation India.}}
  \address{
  $^1$ Computer Science and Engineering, Indian Institute of Technology (IIT), Bombay, India \\
  $^2$ Electrical Engineering, Indian Institute of Science (IISc), Bangalore 560012, India \\
  $^3$ Microsoft Research India, Hyderabad, India \\
  $^4$ Navana Tech India Private Limited, Bangalore, India \\
  $^5$ IBM Research India, Bangalore, India \\
  $^6$ Language Technologies Research Center (LTRC), IIIT Hyderabad, 500032, India
  }
\email{is21\_ss\_indicasr@iisc.ac.in}

\begin{document}
\maketitle
\begin{abstract}
   Recently, there is increasing interest in multilingual automatic speech recognition (ASR) where a speech recognition system caters to multiple low resource languages by taking advantage of low amounts of labeled corpora in multiple languages. With multilingualism becoming common in today’s world, there has been increasing interest in code-switching ASR as well. In code-switching, multiple languages are freely interchanged within a single sentence or between sentences. The success of low-resource multilingual and code-switching ASR often depends on the variety of languages in terms of their acoustics, linguistic characteristics as well as amount of data available and how these are carefully considered in building the ASR system. In this challenge, we would like to focus on building multilingual and code-switching ASR systems through two different subtasks related to a total of seven Indian languages , namely Hindi, Marathi, Odia, Tamil, Telugu, Gujarati and Bengali. For this purpose, we provide a total of $\sim$600 hours of transcribed speech data, comprising train and test sets, in these languages including two code-switched language pairs, Hindi-English and Bengali-English. We also provide baseline recipe\footnote{\scriptsize https://github.com/navana-tech/baseline\_recipe\_is21s\_indic\_asr\_challenge} for both the tasks with a WER of 30.73\% and 32.45\% on the test sets of multilingual and code-switching subtasks, respectively.    
\end{abstract}
\noindent\textbf{Index Terms}: Multilingual ASR, Code-switching

\section{Introduction}
India is a country of language continuum, where every few kilometres, the dialect/language changes \cite{knowLanguages}. Various language families or genealogical types have been reported, in which the vast number of Indian languages can be classified, including Austro-Asiatic, Dravidian, Indo-Aryan, Tibeto-Burman and more recently, Tai-Kadai and Great Andamanese \cite{heitzman1995india,langFamilyGroup}. However, there are no boundaries among these language families; rather, languages across different language families share linguistic traits, including retroflex sounds, absence of prepositions and many more resulting in acoustic and linguistic richness. According to the 2001 census, 29 Indian languages have more than a million speakers. Among these, 22 languages have been given the status of official languages by the Government of India \cite{langFamilies,sailor2020multilingual}. Most of these languages are low resource. Many of these languages do not have a written script, and, hence, speech technology solutions, such as automatic speech recognition (ASR), would greatly benefit such communities \cite{datta2020language}. Another common linguistic phenomenon in multilingual societies is code-switching, typically between between an Indian language and (Indian) English. Understanding code-switching patterns in different languages and developing accurate code-switching ASR remain a challenge due to the lack of large code-switched corpora. 

In such resource-constrained settings, the techniques that exploit unique properties and similarities among the Indian languages could help build multilingual and code-switching ASR systems. Prior works have shown that multilingual ASR systems that leverage data from multiple languages could explore common acoustic properties across similar phonemes or graphemes \cite{chen2020darts,datta2020language,tong2019investigation,lin2009study}. This is achieved by gathering a large amount of data from multiple low-resource languages. Also, multilingual ASR strategies are effective in exploiting the code-switching phenomena in the speech of the source languages. However, there is an emphasis on the need for the right choice of the languages for better performance \cite{cui2015multilingual}, as significant variations between the languages could degrade the ASR performance under multilingual scenarios. In such cases, a dedicated monolingual ASR could perform better even with lesser speech data than a multilingual or code-switching ASR \cite{pratap2020massively,miiller2018multilingual,lin2009study}. 

Considering the aforementioned factors, in this challenge, we have selected six Indian languages, namely, Hindi, Marathi, Odia, Telugu, Tamil and Gujarati, for multilingual ASR; and two code-switched language pairs, Hindi-English and Bengali-English, for the task of code-switching ASR. Unlike prior works on multilingual ASR, the languages selected 1) consider the influences of three major language families -- Indo-Aryan, Dravidian and Austro-Asiatic, which influences most of the Indian languages \cite{langFamilies}, 2) cover four demographic regions of India -- East, West, South and North, and, 3) ensure continuum across languages. It is expected that a multilingual ASR built on these languages could be helpful to extend to other low-resource languages \cite{datta2020language}. Further, most of the multilingual ASR in the literature have considered languages other than Indian languages. Works that consider the Indian languages, however, use the data that is either not publicly available or limited in size \cite{sailor2020multilingual,manjunath2018indian,datta2020language,pratap2020massively,liu2019multilingual}. This challenge significantly contributes in this context, as we provide a large corpus compared to the publicly available data for Indian languages. Publicly available corpora for code-switched speech is also a limited resource and this challenge introduces two freely available datasets consisting of Hindi-English and Bengali-English code-switched speech. 



We provide approximately 600 hours of data in six Indian languages: Hindi, Marathi, Odia, Telugu, Tamil, and Gujarati. In addition, we also provide an additional 150 hours of speech data that includes code-switched transcribed speech in two language pairs, Hindi-English and Bengali-English. Speech recordings in different languages come from different domains. For example, the Odia data comes from healthcare, agriculture and financial domains. The Hindi-English and Bengali-English data are drawn from a repository of technical lectures on a diverse range of computer science topics. We release a baseline system that participants can compare their systems with and use as a starting point. For evaluation of the participating teams, we release held-out blind test sets as well.

The challenge comprises two subtasks. \emph{Subtask1} involves building a multilingual ASR system in six languages: Hindi, Marathi, Odia, Telugu, Tamil, and Gujarati. The blind test set comprises recordings from a subset (or all) of these six languages. \emph{Subtask2} involves building a code-switching ASR system separately for Hindi-English and Bengali-English code-switched pairs. The blind test set comprises recordings from these two code-switched language pairs. Baseline systems are developed considering hybrid DNN-HMM models for both the subtasks, as well as an end-to-end model for the  code-switching subtask. Averaged WERs on the test set and blind test set are found to be 30.73\% \& 32.73\% respectively for subtask1 and for subtask2 those are found to be 33.35\% \& 28.52, 29.37\% \& 32.09\% and 28.45\% \& 34.08\% on test \& blind sets with GMM-HMM, TDNN and end-to-end systems, respectively. The next section elaborates on the two subtasks, providing many details related to dataset creation and specific characteristics of data specific to each subtask.

\begin{table*}
\setlength\tabcolsep{1.5pt}
\centering
\begin{tabular}{|c|c|c|c|c|c|c|c|c|c|c|c|c|c|c|c|c|c|c|}
\hline
 & \multicolumn{3}{|c|}{Hindi} & \multicolumn{3}{|c|}{Marathi} & \multicolumn{3}{|c|}{Odia} & \multicolumn{3}{|c|}{Telugu} & \multicolumn{3}{|c|}{Tamil} & \multicolumn{3}{|c|}{Gujarati} \\
 \hline
     & Trn & Tst & Blnd & Trn & Tst & Blnd & Trn & Tst & Blnd & Trn & Tst & Blnd & Trn & Tst & Blnd & Trn & Tst & Blnd \\
     \hline
     Size (hrs) & 95.05 & 5.55 & 5.49 & 93.89  & 5 & 0.67 & 94.54 & 5.49 & 4.66 & 40 & 5 & 4.39 & 40 & 5 & 4.41 & 40 & 5 & 5.26 \\
     \hline
     Ch.comp & 3GP & 3GP & 3GP & 3GP  & 3GP & M4A & M4A & M4A & M4A & PCM & PCM & PCM & PCM & PCM & PCM & PCM & PCM & PCM \\
     \hline
     Uniq sent & 4506 & 386 & 316 & 2543 & 200 & 120 & 820 & 65 & 124 & 34176 & 2997 & 2506 & 30329 & 3060 & 2584 & 20257 & 3069 & 3419 \\
     \hline
     Spkrs & 59 & 19 & 18 & 31 & 31 & - & - & - & - & 464 & 129 & 129 & 448 & 118 & 118 & 94 & 15 & 18 \\
     \hline
     Vocab (words) & 6092 & 1681 & 1359 & 3245 & 547 & 350 & 1584 & 334 & 334 & 43270 & 10859 & 9602 & 50124 & 12279 & 10732 & 39428 & 10482 & 11424 \\
     \hline
\end{tabular}
\caption{Description of data provided for multilingual ASR (train (Trn), test (Tst) and blind test (Blnd) size, channel compression (Ch.comp), number of unique sentences (Uniq sent), number of speakers (Spkrs) and vocabulary size in words (vocab)). The audio files in all six languages consist of single-channel and are encoded in 16-bit with a sampling rate of 8KHz except for train and test set of Telugu, Tamil and Gujarati, at 16KHz.}

\end{table*}\label{tab_Multi}


\section {Data Details of Two Subtasks}

\subsection{Multilingual ASR}


\noindent \textbf{Motivation.} In India, though there are many low resource languages, they share common language properties from a few language families. Multilingual ASR systems could thus be useful to exploit these common properties in order to build an effective ASR system. Keeping this in mind, the data for this challenge is collected from six languages that are influenced by three major language families. We believe that the data from these languages provide enough variability and, at the same time, cover common properties to build ASR for low resource languages. However, it is challenging to build an ASR considering the six languages due to variability in graphemes, phonemes, and acoustics \cite{indian2016indian}. This session could be useful to address these challenges and build robust multilingual ASR systems. Furthermore, the data from the challenge is made public to the research community to  provide opportunities for better multilingual ASR in the future.


\subsubsection{Dataset Description}\label{sec_dataset}
Table \ref{tab_Multi} shows the data details for the Multilingual ASR specific to each language. The audio files of Odia are collected from four districts as a representative of four different dialect regions -- Sambalpur (North-Western Odia), Mayurbhanj (North Eastern Odia), Puri(Central and Standard Odia) and Koraput (Southern Odia). Further, in all the six languages, the percentage of out-of-vocabulary (OOV) between train \& test and train \& blind test are found to be in the range 17.2\% to 32.8\% and  8.4\% to 31.1\%, respectively. Also, the grapheme set in the data follows the Indian language speech sound label set (ILSL12) standard \cite{indian2016indian}. The total number of graphemes are 69, 61, 68, 64, 50 and 65 respectively for Hindi (Hin), Marathi (Mar), Odia (Oda), Telugu (Tel), Tamil (Tam) and Gujarati (Guj), out of which a total number of diacritic marks in the respective languages are 16, 16, 16, 17, 7 and 17.





\subsubsection{Characteristics of the dataset}
The Hin, Mar and Oda data are collected from the respective native speakers in a reading task. For the data collection, the speakers' choice and the text are selected to cover different language variations for better generalizability. The speakers of Hin and Mar belong to a high-literacy group. On the other hand, the speakers of Odia belong to a semi-literate group. The text data of Hindi and Marathi is collected from storybooks. On the other hand, the text data of Odia is collected from Agriculture, Finance and Healthcare domains. In addition to the speaker and text variability, nativity, other  specific variabilities also exist in the speech data. These variabilities include phoneme mispronunciations and accent variations. In order to retain these variabilities, in the data validation, Oda data went through a manual check. Since Hin and Mar speech data are from a high-literacy group, we consider a semi-automatic process using ASR pipeline for the validation to retain the variabilities.


The automatic validation is done separately for Hindi and Marathi considering the following two measures -- 1) WER obtained from ASR \cite{gretter2014euronews} and 2) likelihood scores obtained from decoded lattice from ASR. In both cases, ASR is trained separately for Hin and Mar, considering noisy data from each language (before validation) of $\sim$ 500hrs. The WER based data validation has been used in prior work \cite{gretter2014euronews}. We believe that the WER-based criteria discards the audios containing insertions, deletions, and/or substitution errors while reading the stimuli. On the other hand, unlike prior works, we include a lattice-based likelihood criteria for discarding very noisy audios or the audios with many incorrect pronunciations. In this work, we consider all those audios whose lattice likelihoods are above 3.49 and WER is equal to 0.0\% for Hindi. Similarly, those audios were chosen for the Marathi data for which the WER is less than or equal to 12.5\% and lattice likelihood is greater than 3.23. The choice of threshold is found to achieve $\sim$ 100hrs of data separately for Hin and Mar by ensuring higher likelihood and lower WERs in the selected audios. Further, the selected data is split into a train (Trn) and test (Tst) without sentence overlap and with out-of-vocabulary (OOV) rates at about 30\% between train and test sets. The Tel, Tam and Guj data are taken from Interspeech 2018 low resource automatic speech recognition challenge for Indian languages, for which, the data was provided by SpeechOcean.com and Microsoft \cite{srivastava2018interspeech}. The train and test sets are considered as-is for this challenge, however, the blind test set is modified with speed perturbations randomly between 1.1 to 1.4 (with increments of 0.05), and/or adding one noise randomly from white, babble and three noises chosen in the Musan dataset \cite{snyder2015musan} considering the signal-to-noise ratio randomly between 18dB to 30dB at step of 1dB. This modification is done randomly on 29.0\%, 23.8\% and 34.1\% of Tel, Tam and Guj data respectively.

\subsection{Code-switching ASR}

\noindent \textbf{Motivation.} Code-switched speech in Indian languages, in the form of publicly available corpora, is a rare resource. This subtask is a first step towards addressing this gap for research on code-switched speech. The code-switched speech is drawn from spoken tutorials on various topics in computer science. These spoken tutorials are available in a wide range of topics and in multiple Indian languages. The tutorials are also accompanied by sentence-level transcriptions and corresponding timestamps. However, this data comes with a number of challenges including label noise which we describe in more detail in Section~\ref{sec:cs-chars}. Understanding these challenges will be important in scaling up the solutions to more real-life data for a larger number of Indian languages.

\begin{table}
\setlength\tabcolsep{1.5pt}
\centering
\begin{tabular}{|c|c|c|c|c|c|c|}
\hline
 & \multicolumn{3}{|c|}{Hindi-English} & \multicolumn{3}{|c|}{Bengali-English}  \\
 \hline
     & Trn & Tst & Blnd & Trn & Tst & Blnd \\
     \hline
     Size (hrs) & 89.86 & 5.18 & 6.24 & 46.11 & 7.02 & 5.53 \\
     \hline
     Uniq sent & 44249 & 2890 & 3831 & 22386 & 3968 & 2936 \\ 
     \hline
     Spkrs & 520 & 30 & 35 & 267 & 40 & 32\\
     \hline
     Vocab (words) & 17830 & 3212 & 3527 & 13645 & 4500 & 3742  \\
     \hline
\end{tabular}
\caption{Description of data provided for code-switching ASR (train (Trn), test (Tst) and blind test (Blnd) size, number of unique sentences (Uniq sent), number of speakers (Spkrs) and vocabulary size in words (vocab))}
\end{table}\label{tab_cs}

\subsubsection{Dataset Description}

The Hindi-English and Bengali-English datasets are extracted from spoken tutorials. These tutorials cover a range of technical topics and the code-switching predominantly arises from the technical content of the lectures. The segments file in the baseline recipe provides sentence time-stamps. These time-stamps were used to derive segments from the audio file to be aligned with the transcripts given in the text file. Table \ref{tab_cs} shows the details of the data considered for code-switched ASR subtasks. All the audio files in both datasets are sampled at 16 kHz, 16 bits encoding. The test-train overlap in Hindi-English and Bengali-English subtasks are 33.9\% and 10.8\% whereas the blindtest-train overlaps are 2.1\% and 2.9\% respectively. Speaker information for both these datasets were not available. However, we do have information about the underlying tutorials from which each sentence is derived. We assumed that each tutorial comes from a different speaker; these are the numbers reported in Table~\ref{tab_cs}. The percentage of OOV words encountered in test and blind-test for Hindi-English subtask is 12.5\% \& 19.6\% and for Bengali-English is 22.9\% \& 27.3\% respectively.


\subsubsection{Characteristics and Artefacts in the Dataset}
\label{sec:cs-chars}

As mentioned earlier, the code-switched speech is drawn from tutorials on various topics in computer science, with transcriptions including mathematical symbols and other technical content. We note here that these tutorials were not created specifically for ASR, but for end-user consumption as videos of tutorials in various Indian languages; specifically in our case, the transcriptions were scripts for video narrators. There are various sources of noise in the transcriptions that are outlined in detail below:
\vspace{0.5em}

\noindent \textbf{Misalignments:} Each spoken tutorial came with transcriptions in the form of subtitles with corresponding timestamps. These timestamps were more aligned with how the tutorial videos proceeded, rather than with the underlying speech signal. This led to misalignments between the transcription and segment start and end times specified for each transcription. While this is an issue for training and development segments, transcriptions corresponding to the blind test segments were manually edited to exactly match the underlyinto remove valid Hindi words and fix any transliteration errors.
\vspace{0.5em}

\noindent \textbf{Inconsistent script usage:} There were multiple instances of the same English word appearing both in the Latin script and the native scripts of Hindi and Bengali in the training data. Given this inconsistency in script usage for English words, the ASR predictions of English words could either be in the native script or in the Latin script. To allow for both English words and their transliterations in the respective native scripts to be counted as correct during the final word error rate computations, we introduce a transliterated WER (T-WER) metric along with the standard WER metric. While the standard WER will only count an edit as correct if it is an exact match with the word in the reference text, T-WER wil count an English word in the reference text as being correctly predicted if it is in English or in its transliterated form in the native script. We compute T-WER rates for the blind test audio files. To support this computation, the blind test reference text was manually annotated such that every English word only appeared in the Latin script. Following this, every English word in the reference transcriptions was transliterated using Google's transliteration API and further manually edited to remove valid Hindi words and fix any transliteration errors. This yielded a list of English to native script mappings and this mapping file was used in the final T-WER to map English words to their transliterated forms.
\vspace{0.5em}

\noindent \textbf{Punctuations:} Since our data is sourced from spoken tutorials on coding and computer science topics, there are many punctuations (e.g., semicolon, etc.) that are enunciated in the speech. This is quite unique to this dataset. There were also many instances of enunciated punctuations not occurring in the text as symbols but rather as words (e.g., 'slash' instead of '/', etc.). There were many non-enunciated punctuations in the text too (exclamation, comma, etc.). We manually edited the blind test transcriptions to remove non-enunciated punctuations and normalized punctuations written as words into their respective symbols (e.g. + instead of plus).
\vspace{0.5em}

\noindent \textbf{Mixed words:} In the Bengali-English transcriptions, we saw multiple occurrences of mixed Bengali-English words (which was unique to Bengali-English and did not show up in Hindi-English). To avoid any ambiguities with evaluating such words, we transliterated these mixed words entirely into Bengali.
\vspace{0.5em}

\noindent \textbf{Incomplete audio:} Since the lecture audios are spliced into individual segments, sometimes a few of the segments have incomplete audio either at the start or at the end of the utterances. For the blind test audio files which went through a careful manual annotation, such words were only transcribed if the word was mostly clear to the annotator. Else, it was omitted from the transcription.
\vspace{0.5em}

\noindent \textbf{Merged English words:} Since the code-switching arises mostly due to technical content in the audio, there are instances of English words being merged together in the transcriptions without any word boundary markers. For example, words denoting websites and function calls generally have two or more words used together without spaces. For the blind test transcriptions, we segregated such merged English words into their constituents in order to avoid ambiguous occurrences of both merged and individual words arising together.

\section {Experiments and Results}

\subsection{Experimental setup}


\subsubsection{Multilingual ASR}

\noindent \textbf{Hybrid DNN-HMM:} ASR model is built using the Kaldi toolkit with a sequence-trained time-delay neural network (TDNN) architecture optimized using the lattice-free MMI objective function \cite{povey2016purely}. We consider an architecture comprising 6 TDNN blocks with dimensionality of size 512.

\noindent \textbf{Lexicon:} A single lexicon is used containing the combined vocabulary of all six languages. For each language, the lexicon's entries are obtained automatically, considering a rule-based system that maps graphemes to phonemes. For the mapping, we consider the Indian speech sound label set (ILSL2) \cite{indian2016indian}. 

\noindent \textbf{Language model (LM)}: A single LM is built considering the text transcriptions belonging to the train set from all six languages. For the LM, we consider a 3-gram language model developed in Kaldi using the IRSTLM toolkit. Since the LM has paths that contain multiple languages, the decoded output could result in code-mixing across the six languages.

In the experiments, word error rate (WER) is considered as the measure for comparison of multilingual ASR. Further for the comparison, we build monolingual ASR systems considering language-specific training data, lexicons and LMs built with language specific train text transcriptions. However, for the evaluation on the blind test, we consider two measures separately to account the channel matching and mismatching scenarios between train/test and blind test. One is averaged WER across all six languages and the other one is averaged WER across all six except Marathi, which has mismatch in the channel encoding scheme.

\subsubsection{Code-switching ASR}

\noindent \textbf{Hybrid DNN-HMM:} The ASR model is built using the Kaldi toolkit, the same baseline architecture being used for both Hindi-English and Bengali-English subtasks. We use MFCC acoustic features to build speaker-adapted GMM-HMM models and similar to  subtask1, we also build hybrid DNN-HMM ASR systems using TDNNs comprising 8 TDNN blocks with dimension 768.

\noindent \textbf{End-to-end ASR:} The hybrid CTC-attention model based on Transformer ~\cite{vaswani2017attention} uses a CTC weight of $0.3$ and an attention weight of $0.7$.  A $12$-layer encoder network, and a $6$-layer decoder network is used, each with $2048$ units, with a $0.1$ dropout rate. Each layer contains eight $64$-dimensional attention heads which are concatenated to form a $512$-dimensional attention vector. Models were trained for a maximum of $40$ epochs with an early-stopping patience of $3$ using the Noam optimizer from~\cite{vaswani2017attention} with a learning rate of $10$ and $25000$ warmup steps. Label smoothing and preprocessing using spectral augmentation is also used. The top 5 models with the best validation accuracy are averaged and this averaged checkpoint is used for decoding. Decoding is performed with a beam size of $10$ and a CTC weight of $0.4$.  

\noindent \textbf{Lexicon:} Two different lexicon files each for Hindi-English and Bengali-English subtasks are used. First, the set of all the words in the training set i.e. the training vocabulary is generated. If the word is a Devanagari/Bengali script word, the corresponding pronunciation is simply the word split into characters. This is because both languages have phonetic orthographies. For English lexicon mappings, an open source g2p package (\url{https://github.com/Kyubyong/g2p}) is used. This spells out numbers, looks up the CMUDict dictionary~\cite{weide1998cmu}, and predicts pronunciations for OOVs as well. We also map punctuations to their corresponding English words since these punctuations are enunciated in the audio.

\noindent \textbf{Language model:} Two separate language models are built for each of the code switching subtasks. For LM training, we consider a trigram language model with Kneser-Ney discounting using the SRILM toolkit developed in Kaldi~\cite{stolcke2002srilm}. 

In the experiments for code-switched speech, along with the standard WER measure, we also provide T-WER values (where T-WER was defined earlier in Section~\ref{sec:cs-chars}). 

\subsection{Baseline results}


\subsubsection{Multilingual ASR}

\noindent \textbf{Comparing multilingual and monolingual ASR:} Table \ref{tab_multiVmono} shows the WERs obtained for test and blind test sets for each of the six languages along with averaged WER across all six languages. We show that the WER obtained with multilingual ASR is lower for Tamil. Though the WER from the multilingual ASR system is higher in the remaining languages, it does not require any explicit language identification system. Thus, the performance of monolingual ASR is affected by the effectiveness of LID for the Indian context. Further, it is known that multilingual ASR is effective in obtaining a better acoustic model by exploring common properties among the multiple languages. However, the performance of the multilingual ASR also depends on the quality of the language model, which, in this work, could introduce noise due to code-mixing of words.

\begin{table}
\setlength\tabcolsep{1.5pt}
\centering
    \begin{tabular}{|c|c|c|c|c|c|c|c|c|}
    \hline
    & & Hindi & Marathi & Odia & Tamil & Telugu & Gujarati & Avg \\
    \hline
    \multirow{2}{*}{Multi}& Tst & 40.41 & 22.44 & 39.06 & 33.35 & 30.62 & 19.27 & 30.73 \\
     & Blnd & 37.20 & 29.04 & 38.46 & 34.09 & 31.44 & 26.15 & 32.73\\
    \hline
    \multirow{2}{*}{Mono} & Tst & 31.39 & 18.61 & 35.36 & 34.78 & 28.71 & 18.23 & 27.85 \\
     & Blnd & 27.45 & 20.41 & 31.28 & 35.82 & 29.35 & 25.98 & 28.38 \\
    \hline
    \end{tabular}
    \caption{Comparison of WER from multilingual and monolingual ASRs on test (Tst) and blind (Blnd) test sets. Averaged WER across five languages on the blind test for Multi and Mono are 33.47 and 29.98 respectively.}
    \label{tab_multiVmono}
\end{table}

\subsubsection{Code-switching ASR}
\begin{table}
\setlength\tabcolsep{1.5pt}
\centering

\begin{tabular}{|c|c|c|c|c|c|c|}
\hline
 &\multicolumn{4}{|c|}{Kaldi-Based} & \multicolumn{2}{|c|}{End-to-End}   \\
\hline
     &\multicolumn{2}{|c|}{GMM-HMM} & \multicolumn{2}{|c|}{TDNN} & \multicolumn{2}{|c|}{Transformer}  \\
 \hline
     & Tst & Blnd & Tst & Blnd & Tst & Blnd   \\
     \hline
     Hin-Eng (UnA) & 44.30 & 25.53 & 36.94  & 28.90 & 27.7 & 33.65  \\
     Ben-Eng (UnA) & 39.19 & 32.81 & 34.31 & 35.52 & 37.2 & 43.94 \\
     \hline
     Avg (UnA) & 41.75 & 29.17 & 35.63 & 32.21  & 32.45 & 38.80 \\
     \hline \hline
     Hin-Eng (ReA) & 31.56 & 24.66 & 28.40 & 29.03 & 25.9 & 31.19  \\
     Ben-Eng (ReA) & 35.14 & 32.39 & 30.34 & 35.15 & 31.0 & 36.97  \\
     \hline
     Avg (ReA) & 33.35 & 28.52  & 29.37 & 32.09  & 28.45 & 34.08  \\
     \hline \hline
\end{tabular}

\caption{WERs from GMM-HMM, Hybrid DNN-HMM and end-to-end ASR systems for Hindi-English (Hin-Eng) and Bengali-English (Ben-Eng) test (Tst) and blind-test (Blnd) sets. (ReA) and (UnA) refers to re-aligned and unaligned audio files, respectively.}
\label{tab_CM_all}
\end{table}

\begin{table}
\setlength\tabcolsep{1.5pt}
\centering

\begin{tabular}{|c|c|c|c|c|c|c|}
\hline
 &\multicolumn{4}{|c|}{Kaldi-Based} & \multicolumn{2}{|c|}{End-to-End}   \\
\hline
     &\multicolumn{2}{|c|}{GMM-HMM} & \multicolumn{2}{|c|}{TDNN} & \multicolumn{2}{|c|}{Transformer}  \\
 \hline
     & WER & T-WER & WER & T-WER & WER & T-WER  \\
     \hline
     Hin-Eng  & 24.66 & 22.72 & 29.03 & 26.20 & 31.19 & 29.80 \\
     \hline
     Ben-Eng  & 32.39 & 31.42  & 35.15 & 33.39 & 36.97 & 36.00 \\
          \hline
     Avg  & 28.52 & 27.07 & 32.09 & 29.79 & 34.08 & 32.9 \\
     \hline
\end{tabular}
\caption{WER and T-WER values for Kaldi and end-to-end based architectures obtained after using aligned (ReA) segments for both Hindi-English (Hin-Eng) and Bengali-English (Ben-Eng) subtasks }
\label{tab_CM_TWER}
\end{table}

Table~\ref{tab_CM_all} shows the WERs (along with the averaged WERs) for both the Hindi-English and Bengali-English datasets.  As mentioned in Section~\ref{sec:cs-chars}, there are misalignments in some of the training audio files between the transcriptions and the timestamps. We present results using  the original alignments that we obtained with the transcriptions (referred to as unaligned, UnA). In an attempt to fix the misalignment issues, we also force-align the training files at the level of the entire tutorial with its complete transcription and recompute the segment timestamps. We retrain our systems using these realigned training files. These numbers are labeled with (ReA) for realigned. 

As expected, we observe that the averaged ReA WERs are consistently better than the UnA WERs. While the Kaldi TDNN-based system gives better WERs for the test set, the speaker adapted triphone GMM-HMM model performs the best among the three ASR systems on the blind test set. 

Table \ref{tab_CM_TWER} shows the corresponding WERs and T-WERs for the realigned (ReA) blind test sets. T-WER, being a more relaxed evaluation metric, is always better than WER. The Hindi-English code mixed data yields improved WERs evidently due to smaller number of OOVs and larger amounts of training data. The corresponding values for the blind-set also improves further as we calculate the transliterated scores as discussed in \ref{sec:cs-chars}.

\section{Conclusion}
In this paper, we present the dataset details and baseline recipe and results for Multilingual and code-switching ASR challenges for low resource Indian languages as a special session in Interspeech 2021. This challenge involves two subtasks dealing with 1) multilingual ASR and 2) code-switching ASR. Through this challenge, the participants have the opportunity to address two important challenges specific to multilingual societies, particularly in the Indian context -- data scarcity and the code-switching phenomena. Through this challenge, we also provide a total of $\sim$600 hours of transcribed speech data, which is reasonably large corpus for six different Indian languages (especially when compared to the existing publicly available datasets for Indian languages). Baseline ASR systems have been developed using hybrid DNN-HMM and end-to-end models. Furthermore, carefully curated held-out blind test sets are also released to evaluate the participating teams' performance.

\bibliographystyle{IEEEtran}

\bibliography{references}

\end{document}